\pdfoutput=1

\documentclass[11pt]{article}

\usepackage[final]{acl}

\usepackage{times}
\usepackage{latexsym}

\usepackage[T1]{fontenc}

\usepackage[utf8]{inputenc}

\usepackage{microtype}

\usepackage{inconsolata}

 \usepackage{amssymb}
\usepackage{amsmath, amsmath}
\usepackage{makecell}
\usepackage{multirow}
\usepackage{array}
\usepackage{booktabs}
\usepackage{soul}
\usepackage[normalem]{ulem}
\usepackage{listings}
\usepackage[switch]{lineno}  %
\usepackage{enumitem}
\usepackage{amssymb}
\usepackage{bbm}
\usepackage[linesnumbered,ruled,vlined]{algorithm2e}

\usepackage{color,xcolor,colortbl}
\usepackage{mathrsfs}
\usepackage{graphicx, float, subfigure}

\newcommand{\ie}{\textit{i}.\textit{e}., }
\newcommand{\eg}{\textit{e}.\textit{g}.\ }

\title{Exploring the Role of Reasoning Structures for Constructing Proofs in Multi-Step Natural Language Reasoning with Large Language Models}

\author{Zi'ou Zheng$^\dagger$, 
~~Christopher Malon$^\ddagger$,
~~Martin Renqiang Min$^\ddagger$,
~~Xiaodan Zhu$^\dagger$\\
$^\dagger$ Department of Electrical and Computer Engineering \\
  \& Ingenuity Labs Research Institute, Queen's University \\ 
  $^\ddagger$NEC Laboratories America \\
  \texttt{\{ziou.zheng,xiaodan.zhu\}@queensu.ca} \quad \\ 
  \texttt{\{malon,renqiang\}@nec-labs.com}
  }

\begin{document}
\maketitle
\begin{abstract}

When performing complex multi-step reasoning tasks, the ability of Large Language Models (LLMs) to derive structured intermediate proof steps is important for ensuring that the models truly perform the desired reasoning and for improving models' explainability. 
This paper is centred around a focused study: whether the current state-of-the-art generalist LLMs can leverage the structures in a few examples to better construct the proof structures with \textit{in-context learning}. Our study specifically focuses on structure-aware demonstration and structure-aware pruning. We demonstrate that they both help improve performance. A detailed analysis is provided to help understand the results.~\footnote{{Our code will be made publicly available at \url{https://github.com/orianna-zzo/structure_reasoning}}} 

\end{abstract}

\section{Introduction}

Large language models (LLMs) have played an essential role in a wide range of applications~\cite{nori2023capabilities,savelka2023explaining,wang-etal-2023-codet5,qin-etal-2023-chatgpt} including intelligent agents~\cite{liu2023agentbench,cheng2022ads}. Their ability to perform complex multi-step reasoning has become  critical~\cite{cot,cot-0shot,tot,besta2023got,lei2023boosting,entbank,street,prontoqa}. In complex multi-hop reasoning tasks, the proof steps often form a graph but not just a chain. The capability to construct correct, structured proofs is essential for ensuring that LLMs perform the desired reasoning and important for the explainability of the reasoning models \cite{entbank,street}. 




In this paper, we perform a focused study, providing evidence to understand whether the state-of-the-art LLMs can leverage a few examples to better construct the proof structure with \textit{in-context learning}. Unlike previous work that fine-tunes the proof models~\cite{hong-etal-2022-metgen,yang-etal-2022-generating,entbank}, we focus on in-context learning~\cite{NEURIPS2020_1457c0d6} since in many applications, the number of available examples with proof structures is small.  
In general, an important goal of generalist models is solving different tasks with the built-in ability and without extensive fine-tuning.

In our research, we consider two key components that can utilize the known proof structures: (i) \textit{demonstration}, and (ii) \textit{proof path search and pruning}. We equip the state-of-the-art LLMs, e.g., \texttt{GPT-4} and \texttt{Llama-3-70B}, with structure-aware demonstration and structure-aware pruning. 
We set up our study with three benchmark datasets, \texttt{EntailmentBank}~\cite{entbank}, \texttt{AR-LSAT}~\cite{street} and \texttt{PrOntoQA}~\cite{prontoqa}. 
The experiment results show that both structure-aware demonstration and structure-aware pruning improve performance. We provide a detailed analysis to help understand the results. 

\begin{figure*}[!t]
\centering
\begin{minipage} [t]{1\linewidth}
\centering
    \includegraphics[width=1\linewidth,clip]{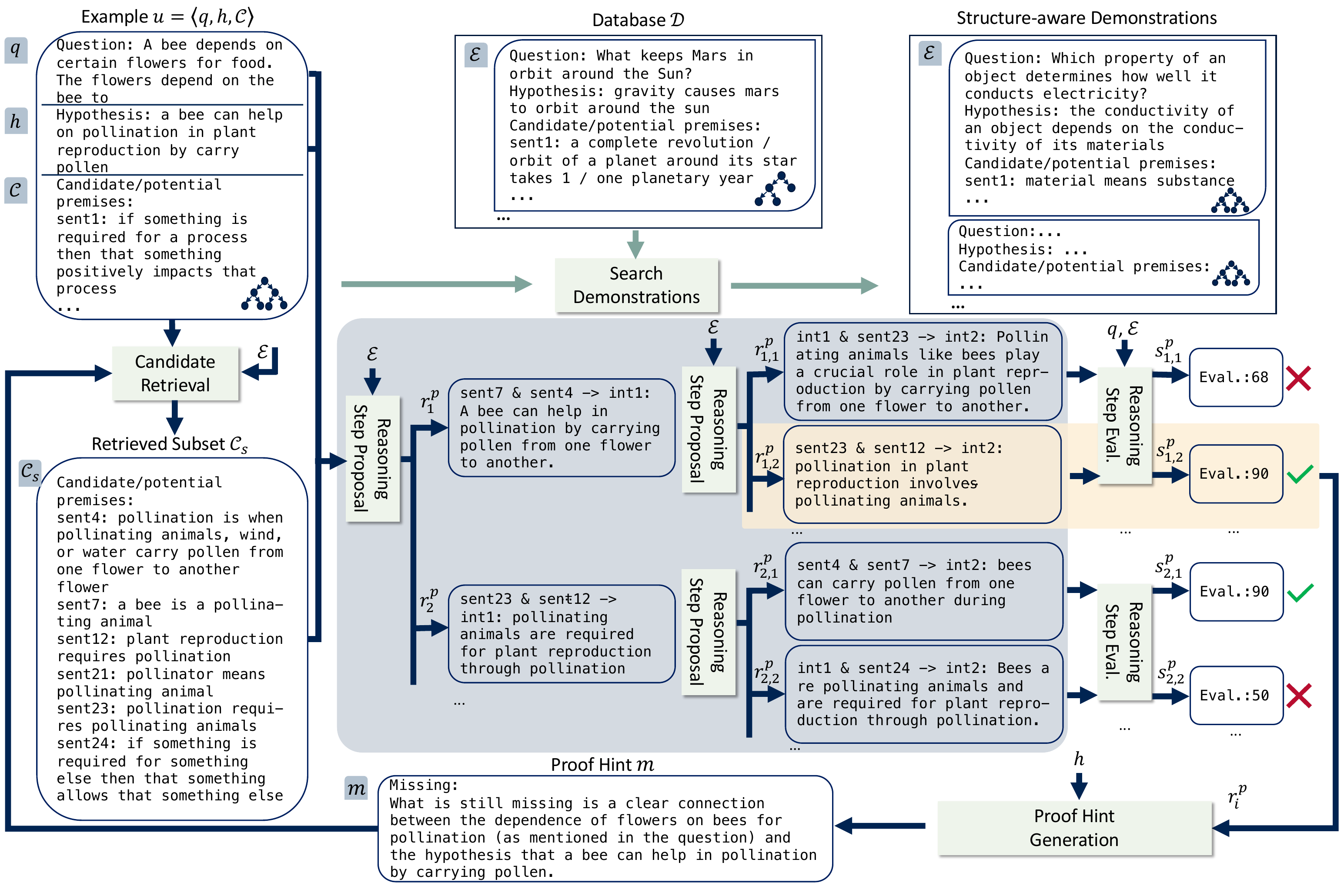}
\end{minipage}
\caption{Overview of each module in our proposed framework. Green arrows indicate the process of searching structure-aware demonstrations, while blue arrows illustrate the proof construction process.}
\label{fig:overview}
\vspace{-2mm}
\end{figure*}

\vspace{-1mm}
\section{Related Work}
\vspace{-1mm}


Recently, LLMs' reasoning ability has been significantly improved. Chain-of-thought (CoT)~\cite{cot,cot-0shot} is arguably the simplest but effective way to elicit linear reasoning chains of LLMs.
Tree-of-thought (ToT)~\cite{tot} can further provide deeper insights into the model's reasoning structures. 
However, ToT has been applied to tasks such as game-of-24 and creative writing, but not to natural language entailment and reasoning tasks with complex proof structures. In this paper, we will compare our models to the CoT and ToT models.

Reasoning in natural language has been a central topic of artificial intelligence research since its inception, including the research in natural language inference~\cite{Dagan2005,snli2015,esim2017,kim2018,feng2022neuro}. In addition to producing accurate results, another key challenge is to improve the explainability of these black-box models, and a variety of recent work has been proposed to address this~\cite{gurrapu2023rationalization,nakamura2023logicattack,zhu2023natlogattack,liu2022a,zini2022explainability,feng2020exploring,valentino2020explainable}.
For complex multi-hop natural language reasoning tasks, 
it is only recently that researchers have begun to develop evaluation datasets to measure proof structure quality~\cite{entbank,street,yang-etal-2022-generating,hong-etal-2022-metgen}. These papers fine-tune small models such as T5~\cite{2020t5} or use older LLMs, lacking studies on structure-aware in-context learning.

\section{Method}
Given a hypothesis $h$ and context $\mathcal{C}$ consisting of evidence sentences, the objective of the task is to provide a proof graph $\mathcal{G}$ to prove $h$ based on some of the evidence sentences in context $\mathcal{C}$, \textit{if} $h$ can be proven. In entailment or multiple-choice question-answering tasks, $h$ is often a concatenation of a question $q$ and a candidate answer, or sometimes a paraphrase of such a concatenation. Since we will use both $q$ and $h$ to retrieve demonstrations and propose reasoning steps, both $q$ and $h$ will be included in our input. As a result, each instance $u$ in the task is a tuple $u=\langle q,h,\mathcal{C} \rangle$.
A proof database $\mathcal{D}$, containing examples of structured proofs, is provided for searching demonstrations that will be used in in-context learning. These demonstrations are exemplars provided to LLMs to help them understand the task requirements and output format.

Formally, we denote $p_\theta$  to be a pre-trained language model with parameter $\theta$. Suppose $x = (x_1, \dots, x_n)$ is a language sequence with $n$ tokens, the probabilistic language model can be written as $p_\theta(x) = \prod_{i=1}^n p_\theta(x_i \mid x_{1, \ldots, i-1})$. Following~\citet{tot}, we use  $p_\theta^{\mathrm{prompt}}(y|x)$ to represent $p_\theta(y|\mathrm{prompt}(x))$ , where $\mathrm{prompt}(x)$ is the input sentences $x$ wrapped with the prompt instructions and templates; $y$ is the output.

In this paper, we hypothesize that the proof structures of similar examples can help LLMs to construct a structured proof for the target problem. In particular, we consider two key components that can utilize the known proof structures: \textit{demonstration} and \textit{proof-path pruning}.
The overall architecture of our model is depicted in Figure~\ref{fig:overview} and Algorithm~\ref{alg:overall}.

At the high level, to support our study, this paper proposes a comprehensive framework consisting of six building blocks: \textit{Structure-aware Demonstration}, \textit{Candidate Retrieval}, \textit{Reasoning Step Proposal}, \textit{Reasoning Step Evaluation}, \textit{Proof Hint Generation}, and \textit{Structure-aware Pruning}, which enables us to perform structure-aware proof construction, and based on that, to conduct a deep study on the advantages and limitations of this construction process within the in-context learning setup.

As shown in Figure~\ref{fig:overview}, given an example $u$ with a question $q$, a hypothesis $h$, and context $\mathcal{C}$, we first derive suitable demonstrations from the provided database $\mathcal{D}$ in the \textit{structure-aware demonstration} stage. Subsequently, the LLM is prompted to retrieve $\mathcal{C}_s$, the subset of the context $\mathcal{C}$, that will be used in the reasoning (\textit{Candidate Retrieval}). Specifically in the example in Figure~\ref{fig:overview}, the LLM returns sentences 4, 7, 12, 21, 23, and 24. With these retrieved sentences, the LLM is asked to provide the next potential reasoning step (\textit{Reasoning Step Proposal}). This can be performed in multiple iterations. For the first iteration, the LLM provides two proof steps: $\mathrm{sent7} \ \& \ \mathrm{sent4}, \mathrm{sent23} \ \& \ \mathrm{sent12}$, and in the second iteration, the LLM proposes two additional steps for each branch. We then evaluate each reasoning step (\textit{Reasoning Step Evaluation}). Next, the model performs structure-guided path pruning and selection (\textit{Structure-aware Pruning}). We note that for each retained branch, the LLM is required to generate a proof hint to guide the next iteration (\textit{Proof Hint Generation}). In the example, we keep two branches and show the generated proof hint for the proposed step $r^p_{1,2}$.

In the following subsections, we will discuss each component in detail and describe how the known proof structures are utilized.

\paragraph{Structure-aware Demonstration.}
Given an example $u=\langle q,h,\mathcal{C} \rangle$ and a database $\mathcal{D}$ where instances feature structured proofs, the search for most similar demonstrations~$\mathcal{E}$ can be expressed as $\mathcal{E} = \mathcal{S}(u, \mathcal{D})$. Usually, $\mathcal{S}$ is defined as manually selecting several fixed demonstrations~\cite{cot,tot} or choosing the top $k$  demonstrations with the example $u$ based on the similarity~\cite{fu2022complexity,liu-etal-2022-makes}.
At the initial stage, we prompt LLMs to provide a guessed proof graph $\mathcal{G}_u^a$ of the example $u$ which is used to find the most similar examples as the demonstrations. As the proof moves forward, the partially constructed proof tree will be simply merged into the guessed tree. 

This process follows the idea of the EM algorithm. We first obtain the guessed structure through LLMs, and then generate one reasoning step based on the demonstrations with the highest similarity after encoding the example with this estimated structure. Subsequently, we update the guessed structure based on the generated reasoning step. The generation of reasoning steps and the updating of the guess structure are performed iteratively.

Specifically, we use the graph attention network (GATv2)~\cite{gatv2} and calculate the similarity between the proof graph $\mathbf{E}_u^a$ and each candidate demonstration $v$'s proof graph $\mathbf{E}_{v}$, which considers both the structure and content of the graphs. We choose the candidates with the higher similarity scores as the demonstrations.


\paragraph{Candidate Retrieval.} 

Given $u=\langle q,h,\mathcal{C} \rangle$, a proof hint $m$ (discussed below), and a set of selected demonstrations $\mathcal{E}$, the candidate retrieval component aims to retrieve a set of most relevant sentences $\mathcal{C}_s$: 
\vspace{-1mm}
\begin{align}
\mathcal{C}_s &= \{o(z_i)\}_{i=1}^k \\
z_i &\sim p_{\theta}^\mathrm{Retrieve} (z|q,h,\mathcal{C},m, \mathcal{E})
\end{align}
where
$z_i$ represents the generated output, which is sampled from the generative language model $p_\theta$ that takes in the retrieval prompt. Because $z_i$ contains the needed sentence \textit{id}, we need to extract the \textit{id} from it; the $o(\cdot)$ represents that extraction process. For detailed examples of prompts, refer to Appendix~\ref{appendix:retrie_prompt}. As a result, $\mathcal{C}_s$ represents a set of retrieved sentences after running the retrieval $k$ times. The proof hint $m$ measures the difference between the
current proof status and the hypothesis, which will be discussed later in the \textit{proof hint generation} subsection. Note that the retrieval models can be replaced by a search engine, but we focus more on reasoning itself.



\paragraph{Reasoning Step Proposal.} 
We then prompt LLMs themselves to provide the most plausible proposal for the next reasoning steps. Formally, given $\langle q, h, \mathcal{C}_s, \mathcal{E} \rangle$, the output is reasoning candidates $r$ for the subsequent reasoning step.
\begin{equation}
    r_i \sim p_{\theta}^\mathrm{Propose} (r|q,h,\mathcal{C}_s, \mathcal{E})
\end{equation}
Then we obtained a set of reasoning steps: $\mathcal{P} = \{r_i\}_{i=1}^{k'}$. 
The output $r_i$ is parsed to transform the output text into a structured step $r^p_i$ such as $\mathrm{sent} \ i \ \& \ \mathrm{sent} \ j\rightarrow\mathrm{int} \ k$. In Figure~\ref{fig:overview}, we can see one such step is $ \mathrm{sent7} \ \& \ \mathrm{sent4}\rightarrow\mathrm{int1}$, meaning intermediate conclusion $\mathrm{int1}$ is drawn from $\mathrm{sent7}$ and $\mathrm{sent4}$.


\paragraph{Reasoning Step Evaluation.} 

Given the current structured reasoning step candidate $r^p_i$ and selected demonstrations $\mathcal{E}$, an LLM measures how likely this reasoning step can reach the final hypothesis with a score $s_i^p$.
\begin{equation}
    s_i \sim p_{\theta}^\mathrm{Eval} (s_i|r^p_i, \mathcal{E})
\end{equation}
where $s_i$ is the language model output from which the score $s^p_i$ is extracted.


\paragraph{Proof Hint Generation.} 
This component asks LLMs to compare the intermediate conclusion $r^p_i$ with the target hypothesis $h$ to provide \textit{proof hint}:
 \begin{equation}
 m_i \sim p_{\theta}^\mathrm{Compare} (m_i|h, r^p_i)
 \end{equation}
An example is shown at the bottom of  Figure~\ref{fig:overview}.  
As discussed above, this will be used to guide the model to find the most relevant evidence.

\paragraph{Search Algorithm and Structure-aware Pruning.} 
During the forward proving process, we combine the typical breadth-first search (BFS) with beam search.
We maintain $b$ beams of candidates, selecting those with the highest evaluation score from the \textit{Reasoning Evaluation} for each exploration. 
Furthermore, we delve into the utilization of the problem's structure in this stage. To explore the effect of structure-guiding path selection, we conducted different experiments on how the structures may be used. In our probing experiment (Appendix~\ref{appendix:preliminary}) on the dev set of EntailmentBank, we found that models benefit from selecting diverse candidate proof steps; \ie the models perform better when they are encouraged to select more diverse candidates. That is, two pieces of evidence located on different subtrees are regarded as more diverse than those on the same subtree. 

Inspired by this, we discourage the model from using the intermediate conclusions which have been used in the previous steps, to avoid growing the tree from the evidence node that has just been generated. 
Specifically, when the \textit{Reasoning Step Proposal} module proposes multiple one-step proofs (\eg $ \mathrm{sent3} \ \& \ \mathrm{sent4}\rightarrow\mathrm{int2}$ or $ \mathrm{sent3} \ \& \ \mathrm{int1}\rightarrow\mathrm{int2}$), the pruning algorithm will consider the proof structure to encourage the newly proposed steps to grow the proof graph from different branches. If a newly proposed proof step grows the graph from the nodes that have just been generated in the previous time step, this proposal will be pruned at this time step (it may still be proposed and used in the future).
We call this implementation the \textit{diversity} (\texttt{div}) variant, which is used in our final model.

\begin{table*}[th!]
\renewcommand\arraystretch{0.8}
    \centering
    \setlength{\tabcolsep}{5pt}
    \footnotesize
    \begin{tabular}{l|l|ccc|ccc|ccc|ccc}
    \toprule
        \multirow{3}{*}{Dataset}  & \multirow{3}{*}{Method} 
        & \multicolumn{3}{c|}{GPT-3.5} &  \multicolumn{3}{c|}{GPT-4}
        & \multicolumn{3}{c|}{Llama-2-70B} &  \multicolumn{3}{c}{Llama-3-70B}\\
        \cmidrule{3-14}
        & &  Ev-F &  Pr-F & G Sim 
         & Ev-F &  Pr-F & G Sim
         &  Ev-F &  Pr-F & G Sim 
         & Ev-F &  Pr-F & G Sim\\
        \midrule
        \multirow{5}{*}{\makecell[c]{EntBank}} 
        & CoT &  .204 &  .059 & .037
              &  .295 &  .128 & .105
              & .196	&.055	  &.035 
              & .281  & .120 & .100\\
        & CoT-sc &  .210 & .062 & .038
                 &  .303  & .138 & .112
                 &.200		&.059  &.037 
                  & .288  & .127 & .110\\
        & ToT  & .220  & .064 & .051
              &  .318 &  .150 & .140
              & .215 &	.062&	.050 
              &.306	&	.143&	.129 \\
        & RAP & .218 & .063 & .050 
              &  .315  & .145 & .135
              &.211	&	.059&	.039 
              & .304 & .139 & .122 \\
        & Ours   & \textbf{.289} & \textbf{.100} & \textbf{.097}
                & \textbf{.355}  & \textbf{.181} & \textbf{.162}
                & \textbf{.261}	&	\textbf{.085}	&\textbf{.071 }
              &  \textbf{.334}	&\textbf{.170}	&\textbf{.145 }\\

        \midrule
        \multirow{5}{*}{\makecell[c]{AR-LSAT}} 
        & CoT &  .472  & .054 & .007
              &  .507  & .078 & .008
              &	.470	&.045	&.006 
              &	.493		&.078	&.008 \\
        & CoT-sc  & .479  & .055 & .007
                  & .524  & .082 & .008
                  &	.482		&.046	&.006 
                 &	.516		&.081	&.008 \\
        & ToT &  .522  & .058 & .008 
              &  .535  & .080 & .008
              &	.488	&	.050	&.006 
              &	.525	&	.080	&.008  \\
        & RAP &  .519 &  .057 & .008 
              &  .532  & .074 & .007
              &	.482	&	.046	&.006 
               &	.527 & .075  & .007  \\
        & Ours  & \textbf{.585}  &\textbf{.079}  & \textbf{.009}
                 & \textbf{.595}  &\textbf{.093}  & \textbf{.010}
                 &\textbf{.515}&	\textbf{.058}&	\textbf{.007}
        	&\textbf{.585}&	\textbf{.089}&	\textbf{.010} \\
      
      \midrule
        \multirow{5}{*}{\makecell[c]{PrOntoQA}} 
        & CoT &  .792 & .760 & .447
              &   .827 & .806 & .528
              &	.789	&.754	&.446 
              &	.818	&.794	&.518 \\
        & CoT-sc & .796 & .765 & .449
                  & .832  & .813 & .530
                  &	.789	&.754	&.446
                &	.828	&.805	&.528\\
        & ToT &  .814  & .779 & .482 
              &   .837 &  .812 & .530
              &.793		&.755	&.447 
        	&.829		&.804	&.528 \\
        & RAP &  .825 &  .786 & .487 
              &   .843  & .820 & .532
              &.802	&.760	&.448 
              & .834 & .810  & .530\\
        & Ours  & \textbf{.837} & \textbf{.798} & \textbf{.504}
               &  \textbf{.852}  & \textbf{.826} & \textbf{.533}
               &	\textbf{.807}	&\textbf{.767}	&\textbf{.450}
                &	\textbf{.844}	&\textbf{.817}	&\textbf{.531}\\
     
     \bottomrule
    \end{tabular}
    \caption{Performance of different models on three benchmark datasets. 
    }  
    \label{tab:res1}
\end{table*}

\section{Experiment Set-Up}

\paragraph{Datasets.} 
We perform experiments on three benchmark datasets, \texttt{EntailmentBank}~\cite{entbank}, \texttt{AR-LSAT}~\cite{street} and \texttt{PrOntoQA}~\cite{prontoqa}. 
Details can be found in Appendix~\ref{appendix:dataset}.
\paragraph{Evaluation Metrics.}
We evaluate the predicted proof graph $\mathcal{G}_{\text{p}}$ against the golden graph $\mathcal{G}_{\text{g}}$ using the following metrics: F1 over evidence~(\texttt{Ev-F})~\cite{entbank}, F1 over proof~(\texttt{Pr-F})~\cite{entbank}, 
and reasoning \texttt{Graph Similarity}~(\texttt{G Sim})~\cite{street}. Details can be found in Appendix~\ref{appendix:evaluation}.

\paragraph{Baselines.}
We compare the proposed method with CoT, self-consistency CoT (CoT-sc), ToT, and reasoning-via-planing (RAP)~\cite{rap}. Each model is prompted with three demonstrations.
Details can be found in Appendix~\ref{appendix:inplement}.
\section{Experiment Results}

We conducted experiments on the representative closed-source (GPT-3.5/GPT-4) and open-source LLMs (Llama-2-70B/Llama-3-70B). 
Table~\ref{tab:res1} shows that our models outperform baseline models across the three datasets under different evaluation metrics. The improvements of the proposed model over baselines are statistically significant (p < 0.05) under one-tailed paired t-test. 
Note that the improvement is less in PrOntoQA, which is due to the fact that a larger percentage of data in PrOntoQA has linear reasoning patterns. Detailed results are included in Appendix~\ref{appendix:full_result}.


\paragraph{Effect of Proof Structure.}
To further understand the effect of proof structures of given examples, we conduct more experiments on EntailmentBank.  Table~\ref{tab:ablation} shows the effectiveness of different components of our model. 
Particularly, our focus is on the variants without structure-aware pruning (``\texttt{w/o prun.}'') and without structure-aware demonstration (``\texttt{w/o demon.}''). We can see that the structure information contributes to the performance (\texttt{Ev-F} and \texttt{G Sim} scores dropped without them.). The comparison involving other variants of our model, specifically concerning the hint module and pruning strategies, is detailed in Table~\ref{tab:ablation_rest} in the Appendix.
The bottom part of Table~\ref{tab:ablation} focuses on evaluating the impact of structure-aware demonstration. We compare the structure-aware demonstration (Ours) vs. regular structure-unaware simple demonstration (Ours$_\text{sim}$). We can see that our model is better under GPT-4. The \texttt{oracle} model means we suppose that we know in advance the proof structure of the question under study and use that to select the most similar demonstrations. 
We can see that our model is effective as its gap from the \texttt{oracle} is not large.




\begin{table}[t!]
\renewcommand\arraystretch{0.8}
    \centering
    \setlength{\tabcolsep}{2pt}
    \footnotesize
    \resizebox{1\linewidth}{!}{
    \begin{tabular}{l|ccccccc}
    \toprule
        Method & Ev-P & Ev-R & Ev-F & Pr-P & Pr-R & Pr-F & G Sim\\
        \midrule
         
         Ours & .388 & .327 & .355 & .204 & .162 & .181 & .162\\
         - w/o prun. & .382 & .311 & .343 & .192 & .159 & .174 & .158\\
         - w/o demon. & .341 & .257 & .293 & .145 & .103 & .120 & .110\\
         - w/o hint &  .339 & .223 & .269 & .140 & .088 & .108 & .093\\
         - w/o retrieval &  .331 & .201 & .250 & .121 & .057 & .077 & .075\\
         


    \midrule
         Ours (w/o prun.) & .382 & .311 & .343 & .192 & .159 & .174 & .158 \\
         Ours$_\text{sim}$ (w/o prun.) & .367 & .258 & .303 & .149 & .121 & .134 & .100\\
         Ours$_\text{oracle}$ (w/o prun.) & .419 & .333 & .371 & .240 & .195 & .215& .205\\

     \bottomrule
    \end{tabular}
    }
    \caption{Ablation analysis on GPT-4. 
    }  
    \label{tab:ablation}
\end{table}

\paragraph{Analysis on Sequential and Non-sequential Reasoning.}
The EntailmentBank dataset consists of reasoning problems that only involve sequential reasoning (the ground-truth proof paths of these problems are chains), as well as non-sequential problems. 
Table~\ref{tab:sequense} depicts the detailed analysis of these two sub-types in the testset with GPT-4. We can see that our method and ToT outperform CoT in both sequential and non-sequential reasoning. Between our model and ToT, they have comparable performance on the sequential subset, while our model performs better than ToT on the non-sequential subset. Our model also outperforms ToT at different depths. Specifically, we conducted a one-tailed t-test over different depths and the differences are statistically significant at depth 3-5 (p < 0.05). For depth 6 and 7, the numbers of available samples are too limited to draw any conclusion.
In general, we can see that non-sequential reasoning is more challenging than sequential reasoning for all models, due to its higher demands on proof planning and development. 
The models not only need to explore new potential premises during reasoning but also ensure that the reasoning process remains coherent. Also, the performances of all models decrease on both sequential and non-sequential problems when the depth increases.

\begin{table}[t!]
\renewcommand\arraystretch{0.95}
    \centering
    \setlength{\tabcolsep}{2pt}
    \resizebox{1\linewidth}{!}{
    \begin{tabular}{c|cc|cc|cc|cc|cc|cc}
    \toprule
        \multirow{4}{*}{Dep.} & \multicolumn{6}{|c|}{Sequential} & \multicolumn{6}{|c}{Non-sequential} \\
        \cmidrule{2-13}
         & \multicolumn{2}{|c|}{CoT}    &\multicolumn{2}{|c|}{ToT}    & \multicolumn{2}{|c}{Ours} & \multicolumn{2}{|c|}{CoT}    &\multicolumn{2}{|c|}{ToT}    & \multicolumn{2}{|c}{Ours}\\
        \cmidrule{2-13}
        & Ev-F & Pr-F & Ev-F & Pr-F & Ev-F & Pr-F & Ev-F & Pr-F & Ev-F & Pr-F & Ev-F & Pr-F \\
        
        \midrule
        3 & .333 & .150 & .356 & .157 & .357 & .157 & .250 & .121 & .266 & .149 & .297 & .151 \\
        4 & .195 & .145 & .242 & .128 & .242 & .129 & .141 & .074 & .160 & .091& .181 & .133 \\
        5 & .102 & .010 & .133 & .015 & .135 & .019 & .057 & .005 & .075 & .006 & .100 & .007 \\
        6 & .013 & .001 & .055 & .005 & .059 & .005 & .011 & .001 & .043 & .003 & .050 & .004 \\
        7 & .002 & .000 & .005 & .002 & .006 & .002 & .002 & .000 & .004 & .001 &.005 & .001\\
        
     \bottomrule
    \end{tabular}
    }
    \caption{Results of sequential /non-sequential reasoning. } 
    \label{tab:sequense}
\end{table}

\section{Conclusion}
Enabling LLMs to generate their proof structure is critical for the reliability and explainability of such models. By incorporating structure-aware components into the state-of-the-art LLMs, we demonstrate that  LLMs can benefit from utilizing the given proof structures of similar examples. We find that measuring the gap between the intermediate steps and the final hypothesis can help narrow down the search space and enhance the performance. 
Further analysis of sequential and non-sequential reasoning reveals that our model offers greater advantages in the more complex task of non-sequential reasoning.



\section*{Limitations}
Our proposed method is primarily designed for the natural language reasoning task, especially the task requiring multi-step proof to obtain the final conclusion. We do not test our method on other types of reasoning, \eg mathematical reasoning and our method is only tested on the English reasoning dataset. 

One limitation, as mentioned in the paper, is the increased token usage with the potential reasoning branches exploration since the system uses LLM-as-a-service API. Although we apply the beam search strategy over the graph which needs less exploration compared to the naive breadth-first search, the overall cost is still high. We also leverage LLM in several modules in the system, which increases the total API calls as well. 

Another limitation is that the current system does not consider the negation proof or the conclusion that cannot be reached. The goal of the current system is to design a system that provides better proof. Proof by negation and other kinds of reasoning, \eg conjunction, disjunction and conditionals, could be extended in future work.




\bibliography{emnlp_official}

\appendix

\section{Preliminary Experiments}\label{appendix:preliminary}
We conduct two preliminary experiments on the dev set of EntailmentBank with GPT-3.5. For the \textit{Preliminary Experiment I}, we provide all other proofs except for randomly deleting two pieces of evidence. We conduct three deletion strategies: two missing pieces of evidence are in the same subtree and the same reasoning step, in the same subtree but not the same reasoning step, or in a different subtree. Here, we set the depth of the subtree to 2. Specifically, ``the same subtree and the same reasoning step'' means the two missing pieces of evidence can together form an intermediate conclusion in the proof tree, while ``the same subtree but different reasoning step'' means that the intermediate conclusion from one missing piece of evidence could be combined with the other missing evidence to obtain another intermediate conclusion. ``A different subtree'' means the two missing pieces of evidence are not in the same 2-depth subtree. Results in Table~\ref{tab:pre_1} show that it is easier for the model to find evidence when they are located in a different proving subtree. We further mimic the practical searching scenario in the \textit{Preliminary Experiment II}, where given one chosen reasoning step, \eg $\mathrm{sent}_4 \ \& \ \mathrm{sent}_5\rightarrow\mathrm{int}_1$, and missed two different reasoning step among which one is based on the given intermediate conclusion (\texttt{reuse\_ic}) and the other (\texttt{div}) is not, \eg $\mathrm{sent}_3 \ \& \ \mathrm{int}_1\rightarrow\mathrm{int}_2$ and $\mathrm{sent}_1 \ \& \ \mathrm{sent}_2\rightarrow\mathrm{int}_3$, we ask the model to provide the prediction of the reasoning step. Table~\ref{tab:pre_2} shows that \texttt{div} model outperforms \texttt{reuse\_ic} and thus we apply \texttt{div} in the main experiment.

\begin{table}[th!]
\renewcommand\arraystretch{0.9}
    \centering
    \setlength{\tabcolsep}{2pt}
    \small
    \resizebox{1\linewidth}{!}{
    \begin{tabular}{l|ccc}
    \toprule
        Model & Ev-P & Ev-R & Ev-F\\
        \midrule
         same subtree and same reasoning step & 0.62  & 0.59 & 0.61\\
         same subtree but different reasoning step  & 0.62  & 0.58 & 0.60\\
         different subtree & 0.63  & 0.60 & 0.62\\

     \bottomrule
    \end{tabular}
    }
    \caption{ Result of Preliminary Experiment I
    }  
    \label{tab:pre_1}
\end{table} 

\begin{table}[th!]
\renewcommand\arraystretch{0.9}
    \centering
    \setlength{\tabcolsep}{2pt}
    \small
    \begin{tabular}{l|cccccc}
    \toprule
        Model & Ev-P & Ev-R & Ev-F & Pr-P & Pr-R & Pr-F\\
        \midrule
         
         reuse\_ic & 0.57 & 0.42 & 0.49 & 0.35 & 0.19 & 0.25\\
         div & 0.59  & 0.45 & 0.51 & 0.36 & 0.19 & 0.25\\

     \bottomrule
    \end{tabular}
    \caption{ Result of Preliminary Experiment II
    }  
    \label{tab:pre_2}
\end{table}

\section{Dataset}\label{appendix:dataset}
\paragraph{EntailmentBank}~\cite{entbank} not only lists the supporting textural evidence but also offers a hierarchical tree structure showing how the evidence organized to lead to the hypothesis. In the entailment tree, the supporting evidence is the leaf node, the hypothesis is the root node, and the intermediate conclusions are the internal nods. 
EntailmentBank is also included in the STREET benchmark~\cite{street}. We exclude the cases which only need one reasoning step, \ie proof depth and length equal to 1.

\paragraph{AR-LSAT} is the Analytical Reasoning -Law School Admission Test task from the STREET benchmark~\cite{street}. STREET benchmark is a unified multi-task and multi-domain natural language reasoning and explanation benchmark. Unlike other existing question-answering (QA) datasets,  models are expected to not only answer questions but also produce step-by-step structured explanations describing how premises in the question are used to produce intermediate conclusions that can prove the correctness of a certain answer. 
We only include AR-LSAT in addition to EntailmentBank because the other datasets in STREET focus on math problems or the sequence process prediction which needs different prompts, especially for the comparison module, with those regarding to logic reasoning in this paper.
For QA datasets, we keep the question as the input $q$ and append the question and correct answer as the input hypothesis $h$. 


\paragraph{PrOntoQA}~\cite{prontoqa} is a synthetic question-answering dataset, where each example is generated from a synthetic world model represented in first-order logic. The rules applied during the synthetic generation endow it with extractable structural information. We applied a similar process on this QA dataset as AR-LSAT except that some examples reasoned by negative deduction are removed in this version.

\section{Implementation Details}~\label{appendix:inplement}
We retrieve 5 times independently and take the union set as the result of the retrieval component. For each step, we propose 3 potential reasoning steps at each node and we keep the beam size as 3 in the breadth-first search. 
The number of demonstrations is set to 3 for all few-shot models. The max iteration number is set to 5 times of the max reasoning depth for each dataset.
We conduct the experiments on gpt-3.5-turbo-0613 version of GPT-3.5 and gpt-4-0125-preview version of GPT-4. For GATv2, we train the model with the training set of EntailmentBank.

\paragraph{Details for ToT and RAP}
We made some modifications for ToT and RAP to better adapt the baselines to the structure-aware natural language reasoning task in this paper. For ToT, we make the thought generator output the potential reasoning step and apply the depth-first-search strategy. RAP needs one fact as the original state to perform the reasoning, which does not work on EntailmentBank and AR-LSAT, and thus we changed the initial state to the void sentence on those datasets.


\section{Algorithms}
The overall algorithm for the proposed method is described in Algorithm~\ref{alg:overall}, and searching structure-aware demonstration and structure-aware pruning can be found in Algorithm~\ref{alg:search_demo} and Algorithm~\ref{alg:pruning} respectively. Specifically in Algorithm~\ref{alg:search_demo}, we use GATv2 as the graph encoder for $\mathrm{GNN}(\cdot)$ and compute cosine similarity score for $\mathrm{sim}(\cdot)$.


\setlength{\textfloatsep}{0pt}
\newcommand\mycommfont[1]{\small\textcolor{blue}{#1}}
\SetCommentSty{mycommfont}
\SetKwInput{KwInput}{Input}             
\SetKwInput{KwOutput}{Output}      
\SetKw{KwInit}{Init}
\SetKwInput{KwReturn}{Return}  
\SetKwInput{KwParam}{Param}  
\SetKwRepeat{Do}{do}{while}
\begin{algorithm}[th!]
\SetAlgoLined
\DontPrintSemicolon
\small

\KwInput{An example $u$ consisting of a question $q$, a hypothesis $h$ and a context $\mathcal{C}$: $u=\langle q,h,\mathcal{C} \rangle$ and a database $\mathcal{D}$}
\KwOutput{Proof graph $\mathcal{G}_p$} 
\KwInit{$\mathcal{G}_p = \varnothing$}\\
\KwInit{Proof hint $m = \varnothing$}\\
\KwInit{$states = \varnothing$}\\
\KwInit{$iter = 0$}\\

\Do{True}
{
 $\mathcal{E} = \mathrm{SearchDemonstration}(u, \mathcal{D}, \mathcal{G}_p)$      // Obtain structure-aware demonstrations \\
 $\mathcal{C}_s = \mathrm{CandidateRetrieval} (q,h,\mathcal{C},m, \mathcal{E})$     
 // Retrieve most relevant sentences \\
 $\{r\}  = \mathrm{ReasoningProposal} (q,h,\mathcal{C}_s, \mathcal{E})$     
 // Obtain reasoning step proposals \\
 \ForEach{$r_i \in \{r\}$}
 {
  $s_i^p = \mathrm{Evaluate}(r_i, \mathcal{E})$ // Obtain evaluation score from \textit{Reasoning Step Evaluation} \\
 }
 $\{\langle r, s^p \rangle \}_{\text{pruned}} = \mathrm{Prune}(\{\langle r, s^p \rangle \}, \mathcal{G}_p, states )$ \\
 $iter += 1$ \\
 \ForEach{$\langle r_i, s_i^p \rangle \in \{\langle r, s^p \rangle \}_{\text{pruned}}$}
 {
    $m_i = \mathrm{GenerateHint}(h, r_i)$ // Generate hint for the next iteration \\
    $\mathcal{G}'_p = \mathcal{G}_p \cup \{\langle r_i, s_i \rangle\}$ \\
    
    \uIf{$\mathcal{G}'_p$ reaches $h$}
    {\KwReturn{$\mathcal{G}'_p$}}
    \uElseIf{$iter > $ the maximum iteration number}
    {\textbf{Continue}}
    \Else{
    Add $\langle \mathcal{G}'_p, m_i, r_i, s_i^p, iter \rangle$ to $states$
    }

 }
 \If{$states == \varnothing$}
 {\textbf{Break}}
 Go to next state in $states$ and update $(\mathcal{G}_p, m, iter)$ with the saved value $(\mathcal{G}'_p, m_i, iter)$ in $states$ \\
}

\caption{Overview}\label{alg:overall}
\end{algorithm}

\SetCommentSty{mycommfont}
\SetKwInput{KwInput}{Input}             
\SetKwInput{KwOutput}{Output}      
\SetKw{KwInit}{Init}
\SetKwInput{KwReturn}{Return}  
\SetKwInput{KwParam}{Param}  
\SetKwRepeat{Do}{do}{while}
\begin{algorithm}[th!]
\SetAlgoLined
\DontPrintSemicolon
\small

\KwInput{An example $u=\langle q,h,\mathcal{C} \rangle$, a database $\mathcal{D}$ and current obtained proof graph $\mathcal{G}_p$}
\KwOutput{Demonstrations $\mathcal{E}$} 

$\mathcal{G}_u^a = \mathrm{GuessGraph(\mathcal{G}_p)}$ // Obtain initial Guessed Proof graph $\mathcal{G}_u^a$ \\

$\mathbf{E}_u^a = \mathrm{GNN}(\mathcal{G}_u^a)$ // Encode graph $\mathbf{E}_u^a$ with GNN \\
\ForEach{$d_i \in \mathcal{D}$}
{
 
 $\mathbf{E}_d{}_i = \mathrm{GNN}(\mathcal{G}_d{}_i)$ // Encode $\mathcal{G}_d{}_i$ with GNN \\

 $s_i = \mathrm{Sim} (\mathbf{E}_d{}_i , \mathbf{E}_u^a)$    // Compute similarity between $\mathbf{E}_d{}_i$ and $\mathbf{E}_u^a$ \\
}

\KwReturn{Demonstrations $\{d_i\}$ with the top $k$ $s_i$}
\caption{SearchDemonstration}\label{alg:search_demo}
\end{algorithm}

\SetCommentSty{mycommfont}
\SetKwInput{KwInput}{Input}             
\SetKwInput{KwOutput}{Output}      
\SetKw{KwInit}{Init}
\SetKwInput{KwReturn}{Return}  
\SetKwInput{KwParam}{Param}  
\SetKwRepeat{Do}{do}{while}
\begin{algorithm}[th!]
\SetAlgoLined
\DontPrintSemicolon
\small

\KwInput{Obtained reasoning steps $\{\langle r, s^p \rangle \}$, current obtained proof graph $\mathcal{G}_p$, and active states $states$}
\KwOutput{Pruned reasoning steps $\{\langle r, s^p \rangle \}_{\text{pruned}}$} 
Keep $\langle r_i, s_i^p \rangle$ where $s_i^p \in \text{top}_k \{s^p\}$ \\
\If{$|\mathcal{G}_p| == 1$}
{
    Delete $\langle r_i, s_i^p \rangle$ where $r_i$ generates conclusion based on intermediate conclusion 
}

\KwReturn{$\{\langle r, s^p \rangle \}_{\text{pruned}}$}
\caption{Prune}\label{alg:pruning}
\end{algorithm}

\section{Evaluation Metrics}~\label{appendix:evaluation}
We evaluate the predicted proof graph $\mathcal{G}_{\text{pred}}$ against the golden graph $\mathcal{G}_{\text{gold}}$ with three metrics, describing evidence, proof and graph similarity. Unlike previous work, we target the model's ability to provide correct proofs more than the true or false result.

\paragraph{Evidence.} Following \cite{entbank}, we perform an evaluation over the chosen evidence to check whether the predicted proof graph uses the correct evidence.  Suppose $E_{\text{pred}}$ and $E_{\text{gold}}$ are the selected evidence set for the predicted proof graph $\mathcal{G}_{\text{pred}}$ and the golden graph $\mathcal{G}_{\text{gold}}$ respectively. We compute precision (\texttt{Ev-P}), recall (\texttt{Ev-R}) and F1 (\texttt{Ev-F}) score by comparing $E_{\text{pred}}$ and $E_{\text{gold}}$ and taking average over the examples.

\paragraph{Proof} Following \cite{entbank}, we evaluate over individual reasoning steps to check whether the predicted proof graph is structurally correct.  Suppose $P_{\text{pred}}$ and $P_{\text{gold}}$ are the reasoning step set for the predicted proof graph $\mathcal{G}_{\text{pred}}$ and the golden graph $\mathcal{G}_{\text{gold}}$ respectively. We compute precision (\texttt{Pr-P}), recall (\texttt{Pr-R}) and F1 (\texttt{Pr-F}) score by comparing $P_{\text{pred}}$ and $P_{\text{gold}}$ and taking average over the examples.

\paragraph{Graph Similarity.} Following \cite{street}, we compute the reasoning graph similarity  (\texttt{G Sim}) $\operatorname{sim}\left(\mathcal{G}_p, \mathcal{G}_g\right)$ by comparing the predicted and the golden reasoning graphs through $\delta\left(\mathcal{G}_p, \mathcal{G}_g\right)$ where $\delta$ is a graph edit distance function using insertion, deletion and substitution as elementary edit operator over nodes and edges. This can be computed as
\begin{equation}
\small
\operatorname{sim}\left(\mathcal{G}_p, \mathcal{G}_g\right)= 
1-\left[\frac{\delta\left(\mathcal{G}_p, \mathcal{G}_g\right)}{\max \left(\left|N_p\right|+\left|E_p\right|,\left|N_g\right|+\left|E_g\right|\right)}\right]
\end{equation}


\section{Additional Results}~\label{appendix:full_result}
Table~\ref{tab:res_gpt} and Table~\ref{tab:res_llama} shows the additional results of precision (\texttt{Ev-P}/\texttt{Pr-P}), recall (\texttt{Ev-R}/\texttt{Pr-R}) of evidence and proof on GPT-3.5, GPT-4, Llama-2-70B and Llama-3-70B. 
The improvements of the proposed model over baselines are statistically significant (p < 0.05) under one-tailed paired t-test.
For example, the p-values of Ev-F and Pr-F (our method vs RAP on ProntoQA) are 0.0247 and 0.0312 for GPT-3.5, while the values are 0.0404 and 0.0388 for GPT-4.
Table~\ref{tab:ablation1_full} and Table~\ref{tab:ablation2_full} shows the additional ablation results on GPT-3.5. 
In Table~\ref{tab:ablation1_full} and Table~\ref{tab:ablation}, in \texttt{w/o prun.}, we do not prune the `proof steps’ based on the structure, while in \texttt{w/o demon.}, we use three fixed demonstrations instead. In Table~\ref{tab:ablation2_full} and Table~\ref{tab:ablation}, the `sim’ variant searches demonstrations only based on the context similarity, without being aware of the structure. Table~\ref{tab:sequense_full} shows the sequential/non-sequential reasoning on GPT-4.
Table~\ref{tab:proof_recall1} shows the proportion of examples where the gold proof is a subset of the predicted proof steps, \ie the proportion of examples where the per-proof recall is 1. This is a stricter metric than the \texttt{Pr-F1}, but it is valuable as it provides insight into the ability of generalist LLMs to produce human-thought correct proofs under different models.

Regarding the guessed structure, we manually examined 20 randomly selected examples (The randomly selected cases are 15, 23, 36, 97, 114, 118, 139, 142, 154, 165, 172, 210, 213, 223, 247, 271, 306, 336, 339, 353 in the EntailmentBank). We found that 8 examples could provide the correct structure in the first guess and the correct number was improved to 12 in the last round of guessing. In comparison to our proposed method, the `sim' variant in Table~\ref{tab:ablation} and Table~\ref{tab:ablation2_full} searches demonstrations only based on the context similarity, without being aware of the structure.

\begin{table*}[th!]
\renewcommand\arraystretch{0.95}
    \centering
    \setlength{\tabcolsep}{3pt}
    \footnotesize
    \resizebox{1\linewidth}{!}{
    \begin{tabular}{l|l|ccccccc|ccccccc}
    \toprule
        \multirow{3}{*}{Dataset}  & \multirow{3}{*}{Model} 
        & \multicolumn{7}{c|}{GPT-3.5} &  \multicolumn{7}{c}{GPT-4} \\
        \cmidrule{3-16}
        & & Ev-P & Ev-R & Ev-F & Pr-P & Pr-R & Pr-F & G Sim 
        & Ev-P & Ev-R & Ev-F & Pr-P & Pr-R & Pr-F & G Sim\\
        \midrule
        \multirow{5}{*}{\makecell[c]{EntBank}} 
        & CoT & .283 & .160 & .204 & .092 & .043 & .059 & .037
              & .326 & .270 & .295 & .152 & .110 & .128 & .105\\
        & CoT-sc & .289 & .165 & .210 & .098 & .045 & .062 & .038
                 & .332 & .279 & .303 & .161 & .121 & .138 & .112\\
        & ToT & .302 & .173 & .220 & .104 & .046 & .064 & .051
              & .347 & .293 & .318 & .174 & .132 & .150 & .140\\
        & RAP & .303 & .170 & .218 & .100 & .046 & .063 & .050 
              & .351 & .285 & .315 & .168 & .128 & .145 & .135\\
        & Ours  & \textbf{.374} & \textbf{.236} & \textbf{.289} & \textbf{.118} & \textbf{.087} & \textbf{.100} & \textbf{.097}
               & \textbf{.388} & \textbf{.327} & \textbf{.355} & \textbf{.204} & \textbf{.162} & \textbf{.181} & \textbf{.162}\\

        \midrule
        \multirow{5}{*}{\makecell[c]{AR-LSAT}} 
        & CoT & .482 & .462 & .472 & .077 & .042 & .054 & .007
              & .523 & .492 & .507 & .092 & .068 & .078 & .008\\
        & CoT-sc & .490 & .468 & .479 & .079 & .042 & .055 & .007
                 & .541 & .508 & .524 & .100 & .070 & .082 & .008\\
        & ToT & .537 & .507 & .522 & .083 & .045 & .058 & .008 
              & .562 & .510 & .535 & .111 & .063 & .080 & .008\\
        & RAP & .539 & .501 & .519 & .079 & .045 & .057 & .008 
              & .571 & .498 & .532 & .098 & .060 & .074 & .007\\
        & Ours  & \textbf{.595} & \textbf{.576} & \textbf{.585} & \textbf{.086} & \textbf{.073} &\textbf{.079}  & \textbf{.009}
                & \textbf{.602} & \textbf{.588} & \textbf{.595} & \textbf{.122} & \textbf{.075} &\textbf{.093}  & \textbf{.010}\\
      
      \midrule
        \multirow{5}{*}{\makecell[c]{PrOntoQA}} 
        & CoT & .802 & .782 & .792 & .782 & .740 & .760 & .447
              &  .843 & .811 & .827 & .812 & .800 & .806 & .528\\
        & CoT-sc & .803 & .790 & .796 & .782 & .748 & .765 & .449
                 & .848 & .816 & .832 & .820 & .806 & .813 & .530\\
        & ToT & .828 & .801 & .814 & .802 & .758 & .779 & .482 
              &  .849 & .825 & .837 & .825 & .800 & .812 & .530\\
        & RAP & .840 & .811 & .825 & .811 & .762 & .786 & .487 
              &  .860 & .827 & .843 & .827 & .814 & .820 & .532\\
        & Ours  & \textbf{.857} & \textbf{.817} & \textbf{.837} & \textbf{.821} & \textbf{.776} & \textbf{.798} & \textbf{.504}
               & \textbf{.866} & \textbf{.838} & \textbf{.852} & \textbf{.831} & \textbf{.821} & \textbf{.826} & \textbf{.533}\\

     \bottomrule
    \end{tabular}
    }
    \caption{Performance of different models on GPT-3.5 and GPT-4. 
    }  
    \label{tab:res_gpt}
\end{table*}

\begin{table*}[th!]
\renewcommand\arraystretch{0.95}
    \centering
    \setlength{\tabcolsep}{3pt}
    \footnotesize
    \resizebox{1\linewidth}{!}{
    \begin{tabular}{l|l|ccccccc|ccccccc}
    \toprule
        \multirow{3}{*}{Dataset}  & \multirow{3}{*}{Model} 
        & \multicolumn{7}{c|}{Llama-2-70B} &  \multicolumn{7}{c}{Llama-3-70B} \\
        \cmidrule{3-16}
        & & Ev-P & Ev-R & Ev-F & Pr-P & Pr-R & Pr-F & G Sim 
        & Ev-P & Ev-R & Ev-F & Pr-P & Pr-R & Pr-F & G Sim\\
        \midrule
        \multirow{5}{*}{\makecell[c]{EntBank}} 
        & CoT & .272 &	.153	&.196	&.087	&.040	&.055	 
              &.035 
              & .318 & .252 & .281 & .143 & .103 & .120 & .100\\
        & CoT-sc & .281 &	.155	&.200	&.092	&.043	&.059  &.037 
                 & .322 & .261 & .288 & .149 & .110 & .127 & .110\\
        & ToT & .293	&.170	&.215	&.100	&.045 &	.062&	.050 
              & .335	&.281	&.306	&.166	&.125&	.143&	.129 \\
        & RAP & .288	&.167	&.211	&.094	&.043&	.059&	.039 
              &  .335	& .279	& .304	& .162	& .121	& .139	& 
 .122\\
        
        & Ours & \textbf{.339}&	\textbf{.212}	&\textbf{.261}	&\textbf{.109} &	\textbf{.069}&	\textbf{.085}	&\textbf{.071 }
              &  \textbf{.361}&	\textbf{.311}	&\textbf{.334}	&\textbf{.194} &	\textbf{.151}&	\textbf{.170}	&\textbf{.145 }\\

        \midrule
        \multirow{5}{*}{\makecell[c]{AR-LSAT}} 
        & CoT & .501	&.443&	.470	&.056	&.037	&.045	&.006 
              & .507	&.479&	.493	&.090	&.069	&.078	&.008 \\
        & CoT-sc & .508	&.458 &	.482	&.059	&.038	&.046	&.006 
                 & .535	&.499&	.516	&.101	&.068	&.081	&.008 \\
        & ToT  &.510	&.467&	.488	&.063	&.042&	.050	&.006 
               &.552	&.501&	.525	&.105	&.065&	.080	&.008  \\
        &RAP &.506	&.458&	.482	&.059	&.038&	.046	&.006 
               &	  .557	&  .500	&  .527	& .096	& .061	& .075	& .007\\
        & Ours  & \textbf{.538}&	\textbf{.493}	&\textbf{.515}&	\textbf{.066}&	\textbf{.051}&	\textbf{.058}&	\textbf{.007}
                & \textbf{.590}&	\textbf{.581}	&\textbf{.585}&	\textbf{.114}&	\textbf{.073}&	\textbf{.089}&	\textbf{.010} \\
      
      \midrule
        \multirow{5}{*}{\makecell[c]{PrOntoQA}} 
        & CoT & .805	&.773&	.789	&.780&	.729	&.754	&.446 
              & .833	&.803&	.818	&.798&	.791	&.794	&.518 \\
        & CoT-sc & .805	&.773&	.789	&.780&	.729	&.754	&.446
                 & .847	&.809&	.828	&.807&	.804	&.805	&.528\\
        
        & ToT &.807&	.779	&.793	&.780&	.732	&.755	&.447 
              &.845&	.814	&.829	&.810&	.799	&.804	&.528 \\
        & RAP & .811&	.793	&.802	&.783&	.738	&.760	&.448 
              & 0.85	& .819	& .834	& .818	& .802	& .810	& .530\\
        & Ours  & \textbf{.813} &	\textbf{.802}&	\textbf{.807}	&\textbf{.785}	&\textbf{.749}	&\textbf{.767}	&\textbf{.450}
                & \textbf{.862} &	\textbf{.827}&	\textbf{.844}	&\textbf{.825}	&\textbf{.810}	&\textbf{.817}	&\textbf{.531}\\
    
     \bottomrule
    \end{tabular}
    }
    \caption{Performance of different models on Llama. 
    }  
    \label{tab:res_llama}
\end{table*}

\begin{table}[th!]
\vspace{-1mm}
\renewcommand\arraystretch{0.95}
    \centering
    \setlength{\tabcolsep}{2pt}
    \small
    \resizebox{1\linewidth}{!}{
    \begin{tabular}{l|ccccccc}
    \toprule
        Model & Ev-P & Ev-R & Ev-F & Pr-P & Pr-R & Pr-F & G Sim\\
        \midrule
         Ours & .374 & .236 & .289 & .118 & .087 & .100 & .097\\
         - w/o prun. & .372 & .230 & .284 & .117 & .087 & .100 & .097\\
         - w/o demon. & .332 & .182 & .235 & .107 & .053 & .071 & .067\\
         - w/o hint &  .313 & .167 & .218 & .103 & .049 & .066 & .064\\
         - w/o retrieval &  .311 & .166 & .216 & .092 & .047 & .062 & .058\\
         
         


     \bottomrule
    \end{tabular}
    }
    \caption{Cumulative ablation analysis. 
    }  
    \label{tab:ablation1_full}
\end{table}

\begin{table}[th!]
\renewcommand\arraystretch{0.9}
    \centering
    \setlength{\tabcolsep}{2pt}
    \small
    \resizebox{1\linewidth}{!}{
    \begin{tabular}{l|ccccccc}
    \toprule
        Model & Ev-P & Ev-R & Ev-F & Pr-P & Pr-R & Pr-F & G Sim\\
        \midrule
        Ours (w/o prun.) & .372 & .230 & .284 & .117 & .087 & .100 & .097\\
        
         Ours$_\text{sim}$ (w/o prun.) & .358 & .211 & .266 & .112 & .069 & .085 & .077\\
         Ours$_\text{oracle}$ (w/o prun.) & .392 & .259 & .312 & .153 & .132 & .142& .138\\

         



     \bottomrule
    \end{tabular}
    }
    \caption{Ablation of demonstration methods.}  
    \label{tab:ablation2_full}
   \vspace{2mm}
\end{table}

\begin{table}[th!]
\renewcommand\arraystretch{0.8}
    \centering
    \setlength{\tabcolsep}{2pt}
    \resizebox{1\linewidth}{!}{
    \begin{tabular}{c|cc|cc|cc|cc|cc|cc}
    \toprule
        \multirow{3}{*}{Dep.} & \multicolumn{6}{|c|}{Sequential} & \multicolumn{6}{|c}{Non-sequential} \\
        \cmidrule{2-13}
         & \multicolumn{2}{|c|}{CoT}    &\multicolumn{2}{|c|}{ToT}    & \multicolumn{2}{|c}{Ours} & \multicolumn{2}{|c|}{CoT}    &\multicolumn{2}{|c|}{ToT}    & \multicolumn{2}{|c}{Ours}\\
        \cmidrule{2-13}
        & Ev-F & Pr-F & Ev-F & Pr-F & Ev-F & Pr-F & Ev-F & Pr-F & Ev-F & Pr-F & Ev-F & Pr-F \\
        \midrule
        3 & .328 & .138 & .330 & .143 & .330 & .144 & .238 & .108 & .257 & .129 & .282 & .135 \\
        4 & .189 & .070 & .202 & .104 & .202 & .113 & .132 & .068 & .149 & .077& .175 & .102 \\
        5 & .082 & .003 & .123 & .007 & .125 & .007 & .049 & .002 & .069 & .005 & .093 & .006 \\
        6 & .012 & .000 & .047 & .004 & .047 & .004 & .010 & .000 & .038 & .003 & .045 & .004 \\
        7 & .002 & .000 & .005 & .001 & .006 & .001 & .002 & .000 & .004 & .001 & .005 & .001\\
        
        
     \bottomrule
    \end{tabular}
    }
    \caption{Results of sequential reasoning /non-sequential reasoning. } 
    \label{tab:sequense_full}
   \vspace{2mm}
\end{table}

\begin{table}[th!]
\renewcommand\arraystretch{0.9}
    \centering
    \setlength{\tabcolsep}{2pt}
    \small
    \begin{tabular}{l|ccc}
    \toprule
        Dataset & Method & GPT-3.5	&GPT-4\\
        \midrule
        \multirow{3}{*}{\makecell[c]{EntailmentBank}} & CoT &.018	& .030 \\
        & ToT	& .024	& .041 \\
        & Ours	& .041	& .071 \\
        \midrule
        \multirow{3}{*}{\makecell[c]{AR-LSAT}} & CoT &.008	&.013\\
        &ToT	&.010	&.015\\
        &Ours	&.018	&.030\\
        \midrule
        \multirow{3}{*}{\makecell[c]{PrOntoQA}} & CoT & .210	& .226 \\
        & ToT	& .229	& .236\\
        & Ours	& .238	& .252\\

     \bottomrule
    \end{tabular}
    \caption{Proportion of the examples where the per-proof recall is 1.}  
    \label{tab:proof_recall1}
   \vspace{2mm}
\end{table}

\section{Other Variants}~\label{appendix:ablation}
Table~\ref{tab:ablation_rest} shows the analysis with other variants of our model. The \texttt{reuse\_ic} variant requires the model to reuse the intermediate conclusion generated in the previous iteration in the 2nd iteration's reasoning, while \texttt{div} variant forces the model to explore the reasoning step from the untouched premises. The \texttt{w/o hint} includes all modules except the \textit{proof hint generation} module. We modify the prompt in this module into asking the model what is the next step of reasoning in \texttt{what's next}. Our findings indicate that the \texttt{div} variant has higher performance than the \texttt{reuse\_ic} and \texttt{w/o pruning} variant, showcasing the effectiveness of the structure-aware pruning.

\begin{table}[th!]
\renewcommand\arraystretch{0.9}
    \centering
    \setlength{\tabcolsep}{2pt}
    \small
    \resizebox{1\linewidth}{!}{
    \begin{tabular}{l|ccccccc}
    \toprule
        Model & Ev-P & Ev-R & Ev-F & Pr-P & Pr-R & Pr-F & G Sim\\
        \midrule
        \multicolumn{8}{c}{GPT-3.5} \\
        \midrule

        Ours (w/o hint) & .359&	.220	&.273&	.100	&.057&	.073&	.072\\

         Ours (what's next)& .363	&.221	&.275	&.108	&.077	&.090&	.089\\
         \midrule
         Ours (w/o pruning) & .372 & .230 & .284 & .117 & .087 & .100 & .097\\
         Ours (reuse\_ic) & .363 & .231 & .282 & .117 & .082 & .096 & .095\\
         Ours (div) & .374 & .236 & .289 & .118 & .087 & .100 & .097\\ 
         
         \midrule
        \multicolumn{8}{c}{GPT-4} \\
        \midrule
         Ours (w/o hint) & .371	&.247	&.297	&.136	&.102	&.117&	.101\\

         Ours (what's next)& .379&	.253	&.303&	.158	&.121&	.137&	.121\\
         \midrule
         Ours (w/o pruning) & .382 & .311 & .343 & .192 & .159 & .174 & .158\\
         Ours (reuse\_ic) & .380 & .309 & .341 & .192 & .157 & .173 & .158\\
         Ours (div) & .388 & .327 & .355 & .204 & .162 & .181 & .162\\ 

     \bottomrule
    \end{tabular}
    }
    \caption{Ablation analysis on EntailmentBank. }  
    \label{tab:ablation_rest}
    \vspace{2mm}
\end{table}

\section{Examples}\label{appendix:example}

\paragraph{Proof Hint Generation.}
Table~\ref{tab:case1} shows two examples with generated hints in the first iteration, and we conduct a comparison between the model with or without the \textit{proof hint generation} module. In the first example, both models could make the correct reasoning in the first iteration and the intermediate conclusion finds out that carbon dioxide is required photosynthesis process. Without the \textit{proof hint generation} module, the model could not retrieve the wanted sentences, while with the \textit{proof hint generation} module, the model succeeds in focusing on the missing relationship with `step'. Similarly, in the second example, both models could correctly retrieve sent6. However, with the \textit{proof hint generation} module, the model cares more about what the Earth revolve around, not the moon. The examples show that the \textit{proof hint generation} module explicitly asks the model to think about the missing part between the current intermediate conclusion and the final goal and the model could retrieve relevant information based on this action. 

\paragraph{Structure-aware Demonstration.}
Table~\ref{tab:case2} shows the example with structure-aware demonstrations. For the page limit, we only show the proof structure of one demonstration in the table. We observe that the model is prone to providing the proof that is structurally similar to the proofs given in the demonstration and we attribute the performance improvement brought by structure-aware demonstrations to this observation.


\begin{table*}[th!]
\renewcommand\arraystretch{0.75}
\centering
\resizebox{1\linewidth}{!}{
    \centering
    \begin{tabular}{c|c|c}
         \toprule
         Model  &  Previous Intermediate Conclusion  &  Retrieved Premises  \\
        
         \midrule  
         
         & \multicolumn{2}{l}{Case 1:}\\
         & \multicolumn{2}{l}{Hypothesis: taking in carbon dioxide is a step in the photosynthesis process}\\
         & \multicolumn{2}{l}{sent2: when carbon dioxide in the atmosphere is absorbed by plants , the amount of carbon dioxide in the atmosphere is reduced in the atmosphere}\\
         &  \multicolumn{2}{l}{sent4: if something is required in a process then obtaining that something is a step in that process}\\
         &   \multicolumn{2}{l}{sent8: taking something in is a kind of method for obtaining that something}\\
         & \multicolumn{2}{l}{sent17: a step in a process means a stage in a process}\\
         & \multicolumn{2}{l}{sent25: taking in and releasing means exchanging}\\

         & \multicolumn{2}{l}{
         Generated hint: What is still missing is a direct connection or evidence that "taking in carbon dioxide" is indeed a step or part of the process of photosynthesis, } \\
         & \multicolumn{2}{l}{
         rather than just a raw material involved in it.} \\
         
         & \multicolumn{2}{l}{Expected: sent4 \& (sent9 \& sent19)}\\
         
         \midrule
         w/o hint & sent9 \& sent19 -> int1: Carbon dioxide is a required raw material in the photosynthesis process. &  sent2, sent12, sent17, sent25
 \\
         w/ hint & sent9 \& sent19 -> int1: Carbon dioxide is required as a raw material in the photosynthesis process. &  sent2, \textbf{sent4}, sent8, sent17\\

          \midrule  
         & \multicolumn{2}{l}{Case 2:}\\
         & \multicolumn{2}{l}{Hypothesis: the difference between the earth and the moon is that the moon revolves around a planet}\\
         & \multicolumn{2}{l}{sent1: celestial bodies / celestial objects are found in space}\\
         & \multicolumn{2}{l}{sent3: earth is a kind of planet}\\
         & \multicolumn{2}{l}{sent4: moons / comets / planets are part of the solar system}\\
         & \multicolumn{2}{l}{sent6: the earth revolves around the sun}\\
         & \multicolumn{2}{l}{sent9: the sun is a kind of star}\\
         & \multicolumn{2}{l}{sent10: a moon is a kind of satellite}\\
         & \multicolumn{2}{l}{sent11: revolving around means orbiting}\\
         & \multicolumn{2}{l}{sent23: the moon is earth 's moon}\\
         & \multicolumn{2}{l}{sent24: a celestial body travelling around another celestial body means that celestial body completes a cycle around that other celestial body}\\
         & \multicolumn{2}{l}{
         Generated hint: What is still missing is evidence that explicitly states that the Earth does not revolve around another planet.} \\
         
         & \multicolumn{2}{l}{Expected: (sent6 \& sent9) \& ((sent25 \& sent3)\& sent11)}\\
         
         \midrule
         w/o hint & sent3 \& sent25 -> int1: The Earth and the Moon are both planets, but the Moon orbits the Earth. & sent1, sent4, \textbf{sent6}, sent10, sent23, sent24 \\
         w/ hint & sent3 \& sent25 -> int1: Earth is a planet and the Moon orbits it. & sent3, \textbf{sent6}, \textbf{sent9}, sent10, \textbf{sent11} \\



         \bottomrule
    \end{tabular}
    }
    \caption{2nd iteration of reasoning examples for w/ and w/o proof hint generation module}
    \label{tab:case1}
    \vspace{-1em}
\end{table*}

\begin{table*}[th!]
\renewcommand\arraystretch{0.75}
\centering
\resizebox{1\linewidth}{!}{
    \centering
    \begin{tabular}{c|c|c}
         \toprule
         Model  &  Demonstration Proof  &  Final Proof \\
        
         \midrule

         & \multicolumn{2}{l}{Hypothesis: wood boards are a kind of building material that is made of a renewable natural resource}\\

         & \multicolumn{2}{l}{sent3: wood boards are made of wood }\\
         &  \multicolumn{2}{l}{sent7: wood is a renewable resource}\\
         &   \multicolumn{2}{l}{sent8: a renewable resource is a kind of natural resource}\\
         & \multicolumn{2}{l}{sent17: wood boards can be used to build houses}\\
         & \multicolumn{2}{l}{sent19: a house is a kind of building}\\
         & \multicolumn{2}{l}{sent23: building materials are used to build buildings}\\
         & \multicolumn{2}{l}{Expected: ((sent19 \& sent23) \& sent17) \& ((sent7 \& sent8) \& sent3)}\\

         \midrule
         Text-aware Demonstration & (sent25 \& sent3) \& sent2  & ((sent7 \& sent8) \& sent17) \\
         Structure-aware Demonstration & ((sent26 \& sent3) \& sent1) \& ((sent7 \& sent9) \& sent10)  & ((sent19 \& sent23) \& sent17) \& ((sent7 \& sent8) \& sent3) \\



         \bottomrule
    \end{tabular}
    }
    \caption{Final proof for structure-aware demonstration and demonstration with the most similar context}
    \label{tab:case2}
    \vspace{-1em}
\end{table*}

\section{Computation Cost}
We observe that the cost of experimenting is higher than the baselines. We leverage the language model in several different modules and apply the beam search strategy in the breadth-first search.  We keep $a$ most promising states per step and $b$ beams of candidates with the highest evaluation score for each exploration in the beam search strategy. Although we cut down the total number of explored cases of $n$ reasoning iterations to $a + (n-1) \times b \times a$ from $a + a^2 + a^3 + \dots + a^n$ because of the beam search over the tree, it is still higher than CoT (1) and ToT ($n \times a$). Table~\ref{tab:sequense} shows our benefits on non-sequential reasoning but similar performance with ToT on sequential reasoning. Considering the computation cost, our model might not be a good choice if most data belongs to sequential reasoning.

\section{Example Prompts}
\label{sec:appendix}
We provide three demonstrations in all few-shot models, but we only show one in the example in this section. 

\subsection{Prompt for Candidate Retrieval}~\label{appendix:retrie_prompt}
System: Below, you are given a question, a hypothesis and a set of candidate premises. You are required to select a small set of candidates (at least provide 3 sentences) to deduce the hypothesis. Please only filter out the sentences that you are sure of.    \\ \\
\text{[example]} \\
Question: What keeps Mars in orbit around the Sun? \\
Hypothesis: gravity causes mars to orbit around the sun \\
Candidate/potential premises: \\
sent1: a complete revolution / orbit of a planet around its star takes 1 / one planetary year  \\
sent2: our sun is located at the center of our solar system  \\
sent3: celestial objects are located in outer space  \\
sent4: gravity causes orbits  \\
sent5: orbit is a kind of characteristic  \\
sent6: a star usually is larger than a planet  \\
sent7: revolving around something means orbiting that something  \\
sent8: a satellite orbits a planet  \\
sent9: uranus is a kind of planet  \\
sent10: planets are found in space  \\
sent11: gravity means gravitational pull / gravitational energy / gravitational force / gravitational attraction  \\
sent12: as mass of a planet / of a celestial body increases, the force of gravity on that planet will increase  \\
sent13: the sun is the strongest source of gravity in the solar system  \\
sent14: a galaxy is made of stars  \\
sent15: orbit means orbital path  \\
sent16: can be means able to be  \\
sent17: celestial bodies / celestial objects are found in space  \\
sent18: satellites are found in space  \\
sent19: proxima centauri is a kind of star  \\
sent20: planets in the solar system orbit the sun  \\
sent21: mars is a kind of planet  \\
sent22: venus is a kind of planet  \\
sent23: mars is located in the solar system  \\
sent24: isaac newton discovered the theory of gravity  \\
sent25: a comet is a kind of celestial body \\ \\
Retrieval sentences (at least 3): sent4, sent20, sent21, sent23 \\
Proof: sent20 \& sent4 -> int1: gravity causes the planets in the solar system to orbit the sun; sent21 \& sent23 -> int2: mars is a planet in the solar system; int1 \& int2 -> hypothesis; \\
---- \\
\text{[Question]} \\
Question: A bee depends on certain flowers for food. The flowers depend on the bee to  \\
Hypothesis: a bee can help on pollination in plant reproduction by carry pollen \\
Candidate/potential premises:  \\
sent1: if something is required for a process then that something positively impacts that process   \\
sent2: pollinated means after pollination   \\
sent3: pollinating is a kind of function   \\
sent4: pollination is when pollinating animals, wind, or water carry pollen from one flower to another flower   \\
sent5: if something causes a process then that something is required for that process   \\
sent6: seed dispersal has a positive impact on a plant / a plant's reproduction 
sent7: a bee is a pollinating animal   \\
sent8: flowers sometimes become fruits after pollination   \\
sent9: if a living thing requires something then that something has a positive impact on that living thing   \\
sent10: flowers are a source of fruit  \\
sent11: if something is required then that something must be provided   \\
sent12: plant reproduction requires pollination   \\
sent13: needing something means depending on that something   \\
sent14: to be used for something means to be required by that something   \\
sent15: flowers often have a sweet smell to attract pollinators   \\
sent16: to carry means to transport   \\
sent17: a bird is a pollinating animal   \\
sent18: a flower's purpose is to produce seeds   \\
sent19: when pollen sticks to a hummingbird, that pollen will move to where the hummingbird moves   \\
sent20: plant requires seed dispersal for reproduction   \\
sent21: pollinator means pollinating animal   \\
sent22: seed dispersal is a kind of method of sexual reproduction   \\
sent23: pollination requires pollinating animals   \\
sent24: if something is required for something else then that something allows that something else   \\
sent25: requiring something means needing that something  \\ \\
Retrieval sentences (at least 3):

\subsection{Prompt for Reasoning Step Proposal}
System: Provide me several sentences with the sentence number and one intermediate conclusion that are possible to be used in the next step in this small set. If the deduction reaches the hypothesis, tell me 'Finish'; otherwise please provide the (intermediate) conclusion.  \\  \\
\text{[example]}  \\
Question: What keeps Mars in orbit around the Sun?  \\
Hypothesis: gravity causes mars to orbit around the sun  \\
Candidate/potential premises:  \\  
sent4: gravity causes orbits   \\
sent5: orbit is a kind of characteristic   \\
sent12: as mass of a planet / of a celestial body increases, the force of gravity on that planet will increase   \\
sent20: planets in the solar system orbit the sun   \\
sent21: mars is a kind of planet   \\
sent22: venus is a kind of planet   \\
sent23: mars is located in the solar system   \\
sent24: isaac newton discovered the theory of gravity   \\ \\
Possible Next Reasoning: sent20 \& sent4 -> int1: gravity causes the planets in the solar system to orbit the sun  \\
--- \\
\text{[Question]}  \\
Question: A bee depends on certain flowers for food. The flowers depend on the bee to  \\
Hypothesis: a bee can help on pollination in plant reproduction by carry pollen  \\
Candidate/potential premises:  \\
sent4: pollination is when pollinating animals, wind, or water carry pollen from one flower to another flower   \\
sent7: a bee is a pollinating animal  \\
sent12: plant reproduction requires pollination  \\
sent21: pollinator means pollinating animal   \\
sent23: pollination requires pollinating animals   \\
sent24: if something is required for something else then that something allows that something else   \\   \\
Possible Next Reasoning:

\subsection{Prompt for Reasoning Step Evaluation}
System: Evaluate whether these intermediate conclusions could reach the hypothesis with candidates. Provide me the number of possibilities (0-99) of these intermediate conclusions: Surely: 85-99, Likely: 50-84, Impossible: 0-49 \\ \\
\text{[example]}  \\
Question: What keeps Mars in orbit around the Sun?  \\
Hypothesis: gravity causes mars to orbit around the sun  \\
Candidate/potential premises:  \\
sent4: gravity causes orbits   \\
sent5: orbit is a kind of characteristic   \\
sent12: as mass of a planet / of a celestial body increases , the force of gravity on that planet will increase   \\
sent20: planets in the solar system orbit the sun   \\
sent21: mars is a kind of planet   \\
sent22: venus is a kind of planet   \\
sent23: mars is located in the solar system   \\
sent24: isaac newton discovered the theory of gravity   \\ \\
Possible Next Reasoning: sent20 \& sent4 -> int1: gravity causes the planets in the solar system to orbit the sun  \\
Evaluate: 99  \\
---  \\
\text{[Question]} \\
Question: The body of a fish is covered by scales for \\
Hypothesis: scales are used for protection by fish \\
Candidate/potential premises: \\
sent1: a fish is a kind of scaled animal \\
sent8: scales are a covering around the body of a scaled animal  \\
sent12: scales are used for protection by scaled animals  \\
sent15: protecting is a kind of function 

\subsection{Prompt for Proof Hint Generation}
System: Compare the intermediate conclusion with the hypothesis and the question, and provide me one sentence of what is still missing. \\ \\
\text{[example]}  \\
Question: What keeps Mars in orbit around the Sun?  \\
Hypothesis: gravity causes mars to orbit around the sun  \\
Intermediate Conclusion: int1: gravity causes the planets in the solar system to orbit the sun  \\
Missing: What is missing is to specifically state that Mars is one of the planets in the solar system.
--- \\
\text{[Question]} \\
Question: The body of a fish is covered by scales for \\
Hypothesis: scales are used for protection by fish \\
Intermediate Conclusion: int1: scales cover the body of a fish \\
Missing: 
\end{document}